\pdfoutput=1

\documentclass[11pt]{article}

\usepackage[]{ACL2023}

\usepackage{times}
\usepackage{latexsym}

\usepackage[T1]{fontenc}

\usepackage[utf8]{inputenc}

\usepackage{microtype}

\usepackage{inconsolata}
\usepackage[graphicx]{realboxes}

\usepackage{xurl}
\title{OpenLLM-Ro - Technical Report on Open-source Romanian LLMs}

\author{ \bf
Mihai Masala$^{a,b}$, 
Denis C. Ilie-Ablachim$^b$,
Dragos Corlatescu$^b$,
Miruna Zavelca$^{a,c}$,\\
\bf
Marius Leordeanu$^b$,
Horia Velicu$^d$,
\bf
Marius Popescu$^c$,
Mihai Dascalu$^b$,
Traian Rebedea$^{b,e}$
\\ \\
$^a$Institute for Logic and Data Science, Bucharest, $^b$National University of Science and\\Technology POLITEHNICA Bucharest, Romania, \\ $^c$University of Bucharest, Romania, $^d$BRD Groupe Societe Generale, $^e$NVIDIA
}
\date{}

\begin{document}
\maketitle
\begin{abstract}
In recent years, Large Language Models (LLMs) have achieved almost human-like performance on various tasks. While some LLMs have been trained on multilingual data, most of the training data is in English. Hence, their performance in English greatly exceeds their performance in other languages.
This document presents our approach to training and evaluating the first foundational and chat LLM specialized for Romanian. 

\end{abstract}

\section{Introduction}
\label{sec:introduction}

The Transformer architecture~\cite{vaswani2017attention} has become ubiquitous recently in the Natural Language Processing (NLP) domain, leading to state-of-the-art performance on various tasks, ranging from text classification to text generation. Decoder-only language models such as GPT series~\cite{radford2018improving,radford2019language,brown2020language} exhibit impressive capabilities such as understanding and generating natural language, learning and solving tasks they have not been directly trained on. In the last few years, Large Language Models (LLMs) development has exploded, with a plethora of models available~\cite{achiam2023gpt,chowdhery2023palm,touvron2023llama,hoffmann2022training,taori2023alpaca,jiang2023mistral}.

Training such massive models requires a vast amount of available data (ranging from tens to hundreds of billions of text tokens) besides computational resources. Naturally, most work on such models focuses mainly on English, often excluding other languages. In this work, we focus on building the resources for building an LLM specialized in Romanian, a less-resourced language. By showcasing our approach and releasing our models, we strive to enable further research and exciting applications for Romanian-speaking users.

Our contributions can be summarized as follows:
\begin{itemize}
    \item We present and release\footnote{\url{www.huggingface.co/OpenLLM-Ro/RoLlama2-7b-Base}} the first foundational Romanian LLM (i.e., RoLlama) based on the open-source Llama 2 model~\cite{touvron2023llama}. Our collection of models is grouped under the OpenLLM-Ro initiative. 
    \item We collect and translate a large set of conversation and instructions, resulting in RoLlama2-7b-Instruct\footnote{\url{www.huggingface.co/OpenLLM-Ro/RoLlama2-7b-Instruct}} and RoLlama2-7b-Chat\footnote{\url{www.huggingface.co/OpenLLM-Ro/RoLlama2-7b-Chat}} models capable of following instructions and entertaining multi-turn conversations in Romanian.
    \item We extend and adapt existing testing suites to evaluate the performance of different models on Romanian tasks.
    \item We make our recipe\footnote{\url{www.github.com/OpenLLM-Ro}} (i.e., datasets and code) and models publicly available to encourage further research both for Romanian and other low or less resourced languages.
\end{itemize}

\section{Related Work}
\label{sec:related_work}

With the advent of open source LLMs, especially Llama2~\cite{touvron2023llama} and Mistral~\cite{jiang2023mistral}, non-English specialized alternatives have been trained. ~\citet{luukkonen2023fingpt} have trained a family of Finnish foundational models, ranging from 186M to 176B parameters. Trained on a total of 38B tokens (with a custom tokenizer), their models showcase state-of-the-art performance on a variety of Finish downstream tasks. Similarly, \citet{cui2023efficient} have trained both a foundational and a chat variant of Llama2 for Chinese, while \citet{wang2023mediagpt} worked on a domain-specific (i.e., media) Chinese LLM. Finetuned chat-models starting from Llama2 or Mistral have been developed for Norwegian\footnote{\url{www.huggingface.co/norallm}}, Dutch\footnote{\url{www.huggingface.co/collections/BramVanroy/geitje-7b-ultra-65c1ee010ad80fd1f6a8f208}}, Bulgarian\footnote{\url{www.huggingface.co/INSAIT-Institute}}, and Serbian\footnote{\url{www.huggingface.co/gordicaleksa/YugoGPT}}.

Unlike other language-specific work, we adopt an open-source approach and are also working on improving the foundational model (by employing continual pretraining) before building a chat variant.

\section{Data}
\label{sec:data}

We employ two distinct data sources for training a Romanian LLM: documents for training the foundational model and instructions for the chat variant. The remainder of this section presents each dataset. 

\subsection{CulturaX}

CulturaX~\cite{nguyen2023culturax} represents a collection of documents in 167 languages. The authors combine mC4~\cite{xue-etal-2021-mt5} and OSCAR~\cite{abadji-etal-2022-towards} multilingual datasets using multiple cleaning and deduplication stages to ensure a high-quality corpus. The Romanian part of CulturaX consists of around 40M documents and 40B tokens.



\subsection{Instructions and Conversations}


We collect a large and diverse dataset of conversations and instructions to build the chat model. Since high-quality data in Romanian is very scarce, we resort to translating datasets such as Alpaca (both with GPT3.5\footnote{\url{www.github.com/tatsu-lab/stanford\_alpaca}} and GPT4\footnote{\url{www.github.com/Instruction-Tuning-with-GPT-4/GPT-4-LLM}}), Dolly\footnote{\url{www.huggingface.co/datasets/databricks/databricks-dolly-15k}}, NoRobots\footnote{\url{www.huggingface.co/datasets/HuggingFaceH4/no\_robots}}, GPT-Instruct\footnote{\url{www.github.com/teknium1/GPTeacher}}, Orca\footnote{\url{www.huggingface.co/datasets/Open-Orca/SlimOrca-Dedup}}, Camel\footnote{\url{www.huggingface.co/datasets/camel-ai/math}}\footnote{\url{www.huggingface.co/datasets/camel-ai/chemistry}}\footnote{\url{www.huggingface.co/datasets/camel-ai/biology}}\footnote{\url{www.huggingface.co/datasets/camel-ai/physics}}, UltraChat\footnote{\url{www.huggingface.co/datasets/stingning/ultrachat/tree/main}}, and OpenAssistant~\cite{kopf2023openassistant}\footnote{\url{www.huggingface.co/datasets/OpenAssistant/oasst1}}. Additionally, we manually translated the MTBench corpora \cite{zheng2023judging}, leading to a total of around 2.3M instructions and conversations that were translated\footnote{\url{www.systransoft.com/} last accessed 6th of March 2024} and used for our experiments.

\section{Training}
\label{sec:training}

Training the foundational and chat variants is done using the official Llama repository\footnote{\url{www.github.com/facebookresearch/llama-recipes}}. We pick a fixed learning rate from early experiments with a small portion of the data instead of the cosine scheduler proposed in the original paper. Due to hardware limitations, we experiment with a maximum sequence length of 512, 1024, and 2048, picking the biggest batch size that fits into the memory (i.e., 64 for the sequence length of 512, 32 for the sequence length of 1024, and 16 for the sequence length of 2048). We use the proposed AdamW optimizer, with weight decay set to 0.1 and gradient clipping. 
We do not employ LORA~\cite{hu2021lora} in any stage, as we perform full training for both the continual pre-training and supervised finetuning steps.

\subsection{Foundational Model}

Training the foundational model is done iteratively using continual pretraining on the CulturaX dataset. Starting from the base Llama2\footnote{\url{www.huggingface.co/meta-llama/Llama-2-7b-hf}} model, we train the first model on 5\% of the CulturaX, the following model on the next 5\% (leading to the second model seeing 10\% of the data), and the third model on another 10\% of the CulturaX. Therefore, we have three variants of a Romanian foundational model, variants that used 5\%, 10\%, and 20\% of the entire CulturaX dataset. Training the foundational model is done over one epoch, with a fixed learning rate of \texttt{1e-4}. 

\subsection{Instruct \& Chat Models}
We train RoLlama2-7b-Instruct using translated instructions while we further include the translated conversations from UltraChat and OpenAssistant to the finetuning dataset for RoLlama2-7b-Chat. Training is done with a fixed learning rate of \texttt{5e-5} over one epoch. We additionally train an instruct variant (with no pretraining on Romanian) of Mistral foundational model, leading to RoMistral-7b-Instruct\footnote{\url{www.huggingface.co/OpenLLM-Ro/RoMistral-7b-Instruct}}. Training a RoMistral foundational model coupled with improved Instruct and Chat variants will follow shortly.

\section{Evaluation}
\label{sec:evaluation}
Evaluating a general LLM is not a straightforward task considering the vast number of benchmarks and evaluation protocols proposed to judge the quality of a model. Popular benchmarks include Open LLM Leaderboard\footnote{\url{www.huggingface.co/spaces/HuggingFaceH4/open\_llm\_leaderboard}}, an online leaderboard that encompasses six reasoning and general knowledge tasks: ARC~\cite{clark2018think}, HellaSwag~\cite{zellers2019hellaswag}, MMLU~\cite{hendrycks2020measuring}, TruthfulQA~\cite{lin2021truthfulqa}, Winogrande~\cite{sakaguchi2021winogrande} and GSM8k~\cite{cobbe2021training}. MT-Bench~\cite{zheng2024judging} proposes a multi-turn question set and shows that strong LLMs (e.g., GPT-4) can act as judges and can approximate human preferences in a scalable and somewhat explainable way. AGIEval~\cite{zhong2023agieval} is a benchmark designed to evaluate foundational models in the context of human-centric standardized exams. Evaluation is done in multiple settings, from zero and few-shot to Chain of Thought (COT) prompting. While benchmarks in the Open LLM Leaderboard select the most probable answer (out of four candidates) for each question, the focus on MT-Bench is more on the quality of the generated text than the intrinsic knowledge of the world of each model. Some experiments\footnote{\url{www.twitter.com/maximelabonne/status/1746317053143773628?s=19}} try and compute a correlation factor between human preferences and results of previously mentioned benchmarks.

Evaluating LLMs in any language besides English only adds another layer of complexity. For Romanian versions of ARC, HellaSwag, MMLU, and TruthfulQA, we resort to ChatGPT translations~\cite{dac2023okapi}, while we translate\footnote{same system used for translating instructions} Winogrande and GSM8k. The generally used few-shot values for ARC, HellaSwag, and MMLU (i.e., 25 shots for ARC, 10 shots for HellaSwag, and 5 shots for MMLU) lead to the prompt having a length equal to or even surpassing 2048 tokens. As some of our models have been trained only on sequences of length 512, 1024, or 2048, we opt to use multiple few-shot settings. More specifically, for ARC, we test with 0, 1, 3, 5, 10, and 25 shots, MMLU with 0, 1, 3, and 5, HellaSwag with 0, 1, 3, 5, and 10 shots, and Winogrande is evaluated in 0, 1, 3 and 5 shot setting. For GSM8k, we employ 1, 3, and 5 shot evaluation. Results in Tables~\ref{tab:overview}, \ref{tab:foundational} are averaged over a few shot settings for each benchmark. More detailed results are presented in the Appendix.

\begin{table*}[ht]
\centering
\begin{tabular}{lccccccc}
\hline
\textbf{Model} & \textbf{Average} & \textbf{ARC} & \textbf{MMLU} & \textbf{Wino} & \textbf{HS} &  \textbf{GSM8k} & \textbf{TQA}\\
\hline
\multicolumn{8}{c}{\textbf{\textit{Existing Models}}}\\
\hline
\multicolumn{8}{c}{\textit{For Romanian}}\\
Okapi & 36.05&	36.88	&31.17&	56.28&	48.91&	0.43	&42.64\\
Andrei481/Llama-2-7b-Romanian & 36.53&	35.49&	30.86&	55.98&	47.99&	0.83	&48.01\\
Andrei481/Mistral-7B-v0.1-Romanian &\textbf{37.42}&	37.80	&33.34	&56.37	&49.42&	0.99&	46.63\\
\hline
\multicolumn{8}{c}{\textit{General}}\\
Llama-2-7b & 35.65	&33.85&	30.93&	56.43&	46.98&	1.37	&44.36\\
Llama-2-7b-chat & 35.58&	34.92&	32.37	&54.26&	44.52&	2.05&	45.38\\
Mistral-7B-v0.1 & 42.66&	41.22&	40.84&	60.20&	53.63&	13.82&	46.24\\
Mistral-7B-Instruct-v0.2  & \textbf{45.63}&	43.09&	44.87&	59.26	&54.12	&10.86&	61.56\\
\hline\hline
\multicolumn{8}{c}{\textbf{\textit{Romanian LLMs}}}\\
\textit{RoLlama2-7b-Base} & 38.32&	35.83&	30.47&	60.16	&55.52	&2.17&	45.78\\
\textit{RoLlama2-7b-Instruct} & 44.42	&40.36	&37.41&	69.58	&55.64	&17.59&	45.96\\
\textit{RoLlama2-7b-Chat} & 42.65 & 38.29	&35.27&	65.25	&56.45&12.84&	47.79\\
\hline
\textit{RoMistral-7b-Instruct} & \textbf{52.49} & 50.39	&51.64&	66.69	&60.24&33.71&	52.29\\
\hline
\end{tabular}
\caption{Comparison between RoLLMs and other LLMs on Romanian versions of downstream datasets (abbreviations: HS - HellaSwag, Wino - Winogrande, TQA - TruthfulQA). \textbf{Bold} denotes the best results within a category.}
\label{tab:overview}
\end{table*}

\begin{table*}[t]
\centering
\begin{tabular}{lllccccccc}
\hline
\textbf{Model} & \textbf{Seq. len} & \textbf{\% Data}&  \textbf{Average} & \textbf{ARC} & \textbf{MMLU} & \textbf{Wino} & \textbf{HS} &  \textbf{GSM8k} & \textbf{TQA}\\
\hline
RoLlama2-7b-Base & 512 & 5 & 38.01&	35.07	&32.07&	58.72	&54.80&	3.08&	44.33\\
RoLlama2-7b-Base & 1024 & 5 & \textbf{38.82}&	35.98&	32.10&	59.85	&55.27	&4.22&	45.50\\
RoLlama2-7b-Base & 2048 & 5 & 38.80	&35.98&	32.08	&59.55&	55.07&	4.47&	45.65\\
\hline
RoLlama2-7b-Base & 1024 & 10 & 38.61&	35.75	&31.94&	60.20	&55.58	&3.43	&44.77\\
RoLlama2-7b-Base & 1024 & 20 & 38.32	&35.83&	30.47&	60.16&	55.52&	2.17&	45.78\\
\hline
\end{tabular}
\caption{Foundational models performance under various learning setups. Same notations as in Table \ref{tab:overview}.}
\label{tab:foundational}
\end{table*}

\begin{table*}[h!]
\centering
\begin{tabular}{lccc|c}
\hline
\textbf{Model} & \textbf{Average} & \textbf{1st turn} & \textbf{2nd turn} & \textbf{\#Answers in Ro}\\
\hline
\multicolumn{4}{c}{\textbf{\textit{Existing Models}}}\\
\hline
\multicolumn{4}{c}{\textit{For Romanian}}\\
Okapi & 3.19 & 3.70 & 2.68 & 156 / 160\\
Andrei481/Llama-2-7b-Romanian & 0.82 & 0.99 & 0.65 & 157 / 160 \\
Andrei481/Mistral-7B-v0.1-Romanian & 1.36 & 2.45 & 0.27 & 158 / 160\\
\hline
\multicolumn{4}{c}{\textit{General}}\\
Llama-2-7b-chat & 0.80 & 1.34 & 0.25 & 21 / 160 \\
Llama-2-7b-chat (\textit{ro\_prompted}) & 1.39 & 2.05 & 0.73 & 43 / 160 \\
Mistral-7B-Instruct-v0.2 & 1.09 & 1.40 & 0.78 & 24 / 160 \\
Mistral-7B-Instruct-v0.2 (\textit{ro\_prompted}) & 5.84 & 6.06 & \textbf{5.63} & 151 / 160 \\
\hline\hline
\multicolumn{4}{c}{\textbf{\textit{Romanian LLMs}}}\\
\textit{RoLlama-Instruct} & 4.31 & 5.66 & 2.95 & \textbf{160 / 160}\\
\textit{RoLlama-Chat} & 3.91 & 4.25 & 3.57 & 154 / 160\\
\textit{RoMistral-Instruct} & \textbf{5.92} & \textbf{6.53} & 5.41 & \textbf{160 / 160}\\

\hline
\end{tabular}
\caption{Chat models performance on translated MT-Bench in Romanian. \textbf{Bold} denotes the best result.}
\label{tab:chat-mtbench}
\end{table*}


\section{Results}
\label{sec:related_work}

Table~\ref{tab:overview} introduces the evaluation results for existing LLMs compared to the foundational and chat variants of RoLlama. Results show that training a Romanian LLM improves scores over general LLMs: around 3\% for the foundational models and around 9\% for the chat variant.

We observe a very high performance of Mistral models (both foundational and chat). However, since the authors of Mistral (\citet{jiang2023mistral}) do not provide many details about their training data, we cannot know if the model has been trained on Romanian data as well as other languages. Furthermore, it seems that finetuning Mistral on Romanian Alpaca~\footnote{\url{www.huggingface.co/Andrei481/Mistral-7B-v0.1-Romanian/discussions/1}} drastically decreases the performance on considered benchmarks. 

Turning to an evaluation based more on generation rather than just picking the most probable continuation, the performance on MT-Bench is presented in Table \ref{tab:chat-mtbench}. We note that even when using a Romanian prompt, most Llama2 and Mistral completions or answers are either entirely in English or have the first few words in Romanian, then reverting to English. Specifically asking the models to answer in Romanian in the prompt somewhat alleviates this issue, especially for Mistral models. As seen in the last column in Table \ref{tab:chat-mtbench}, when specifically prompted to answer in Romanian, Llama2 manages to generate text in Romanian in only 25\% of cases, while Mistral answers are almost entirely in Romanian (over 90\% of cases vs 70\% when not prompted to answer in Romanian). We automatically detect the language of the answer\footnote{\url{www.pypi.org/project/langdetect/}} and overwrite with 0 the score of each answers that is not in Romanian.

As expected, our models can mostly generate text in Romanian without being specifically told to do so. The chat model also generates 6 texts in English, which might be explained by the presence of some translation errors in the training conversations. With the exception of the Mistral model, the second-turn scores for all models are much lower than the first-turn scores. For RoLlama2-7b-Instruct, this is somewhat expected as the model has predominantly seen single-turn conversations during training. While still underperforming, the first turn score of RoLlama2-7b-Instruct is within 0.5 points of Mistral. Adding multi-turn conversations (i.e., in the form of UltraChat and Open Assistant) increases the score on second turns, but the overall score is still lower. The same phenomenon can also be observed in Table~\ref{tab:overview}. As the conversations account for about 70\% of the finetuning data (around 1.6M samples), their quality directly affects the performance of the final model. Methods for validating and curating the translations and selecting high-quality samples for building the best data blend should be researched further as they could lead to lower training times and better-performing models.
RoMistral-7b-Instruct further improves the quality of the first turn answer compared with Mistral-Instruct, becoming the best performing model. On second turn answers the difference is small. The Chat variant of RoMistral will probably outperform Mistral-Instruct in the quality of both turns.

\section{Conclusions}
\label{sec:conclusion}

This technical report presents initial experiments performed for building the first open-source LLM specialized for Romanian. While still in their infancy, the released models show promising results, outperforming existing solutions for various tasks. Pretraining on additional cleaner data, alignment with human preference~\cite{ziegler2019fine,rafailov2024direct,dong2023steerlm}, finetuning with adapters~\cite{hu2021lora}, validating existing translations or adding more instructions and conversations are just a few improvements that should increase the quality of RoLlama family of models. Note that we inherit the model and dataset licenses for all variants of RoLlama. Further work should include a more permissive combination of datasets for finetuning the Chat and Instruct variants.

Furthermore, we present a general recipe which we expect to work on different, more powerful LLMs as well (e.g., bigger Llama2, Mistral, Mixtral\footnote{\url{www.mistral.ai/news/mixtral-of-experts/}} or Llama3\footnote{\url{ai.meta.com/blog/meta-llama-3/}} models). We firmly believe that this is just the first step towards building and adapting LLMs for Romanian for both research and industry applications.

\section*{Acknowledgements}

The universities contributed with researchers who worked pro-bono (POLITEHNICA Bucharest and the University of Bucharest) and computing power for training the models (POLITEHNICA Bucharest). Synchronization and collaboration took place within a project\footnote{\url{www.ilds.ro/llm-for-romanian}} carried out in the Institute of Logic and Data Science with funding from BRD Groupe Societe Generale.

We thank Alin Stefanescu and Laurentiu Leustean for participating in various discussions regarding this project.

\bibliographystyle{acl_natbib}
\bibliography{custom}

\begin{thebibliography}{31}
\expandafter\ifx\csname natexlab\endcsname\relax\def\natexlab#1{#1}\fi

\bibitem[{Abadji et~al.(2022)Abadji, Ortiz~Suarez, Romary, and Sagot}]{abadji-etal-2022-towards}
Julien Abadji, Pedro Ortiz~Suarez, Laurent Romary, and Beno{\^\i}t Sagot. 2022.
\newblock \href {https://aclanthology.org/2022.lrec-1.463} {Towards a cleaner document-oriented multilingual crawled corpus}.
\newblock In \emph{Proceedings of the Thirteenth Language Resources and Evaluation Conference}, pages 4344--4355, Marseille, France. European Language Resources Association.

\bibitem[{Achiam et~al.(2023)Achiam, Adler, Agarwal, Ahmad, Akkaya, Aleman, Almeida, Altenschmidt, Altman, Anadkat et~al.}]{achiam2023gpt}
Josh Achiam, Steven Adler, Sandhini Agarwal, Lama Ahmad, Ilge Akkaya, Florencia~Leoni Aleman, Diogo Almeida, Janko Altenschmidt, Sam Altman, Shyamal Anadkat, et~al. 2023.
\newblock Gpt-4 technical report.
\newblock \emph{arXiv preprint arXiv:2303.08774}.

\bibitem[{Brown et~al.(2020)Brown, Mann, Ryder, Subbiah, Kaplan, Dhariwal, Neelakantan, Shyam, Sastry, Askell et~al.}]{brown2020language}
Tom Brown, Benjamin Mann, Nick Ryder, Melanie Subbiah, Jared~D Kaplan, Prafulla Dhariwal, Arvind Neelakantan, Pranav Shyam, Girish Sastry, Amanda Askell, et~al. 2020.
\newblock Language models are few-shot learners.
\newblock \emph{Advances in neural information processing systems}, 33:1877--1901.

\bibitem[{Chowdhery et~al.(2023)Chowdhery, Narang, Devlin, Bosma, Mishra, Roberts, Barham, Chung, Sutton, Gehrmann et~al.}]{chowdhery2023palm}
Aakanksha Chowdhery, Sharan Narang, Jacob Devlin, Maarten Bosma, Gaurav Mishra, Adam Roberts, Paul Barham, Hyung~Won Chung, Charles Sutton, Sebastian Gehrmann, et~al. 2023.
\newblock Palm: Scaling language modeling with pathways.
\newblock \emph{Journal of Machine Learning Research}, 24(240):1--113.

\bibitem[{Clark et~al.(2018)Clark, Cowhey, Etzioni, Khot, Sabharwal, Schoenick, and Tafjord}]{clark2018think}
Peter Clark, Isaac Cowhey, Oren Etzioni, Tushar Khot, Ashish Sabharwal, Carissa Schoenick, and Oyvind Tafjord. 2018.
\newblock Think you have solved question answering? try arc, the ai2 reasoning challenge.
\newblock \emph{arXiv preprint arXiv:1803.05457}.

\bibitem[{Cobbe et~al.(2021)Cobbe, Kosaraju, Bavarian, Chen, Jun, Kaiser, Plappert, Tworek, Hilton, Nakano et~al.}]{cobbe2021training}
Karl Cobbe, Vineet Kosaraju, Mohammad Bavarian, Mark Chen, Heewoo Jun, Lukasz Kaiser, Matthias Plappert, Jerry Tworek, Jacob Hilton, Reiichiro Nakano, et~al. 2021.
\newblock Training verifiers to solve math word problems, 2021.
\newblock \emph{URL https://arxiv. org/abs/2110.14168}.

\bibitem[{Cui et~al.(2023)Cui, Yang, and Yao}]{cui2023efficient}
Yiming Cui, Ziqing Yang, and Xin Yao. 2023.
\newblock Efficient and effective text encoding for chinese llama and alpaca.
\newblock \emph{arXiv preprint arXiv:2304.08177}.

\bibitem[{Dac~Lai et~al.(2023)Dac~Lai, Van~Nguyen, Ngo, Nguyen, Dernoncourt, Rossi, and Nguyen}]{dac2023okapi}
Viet Dac~Lai, Chien Van~Nguyen, Nghia~Trung Ngo, Thuat Nguyen, Franck Dernoncourt, Ryan~A Rossi, and Thien~Huu Nguyen. 2023.
\newblock Okapi: Instruction-tuned large language models in multiple languages with reinforcement learning from human feedback.
\newblock \emph{arXiv e-prints}, pages arXiv--2307.

\bibitem[{Dong et~al.(2023)Dong, Wang, Sreedhar, Wu, and Kuchaiev}]{dong2023steerlm}
Yi~Dong, Zhilin Wang, Makesh~Narsimhan Sreedhar, Xianchao Wu, and Oleksii Kuchaiev. 2023.
\newblock Steerlm: Attribute conditioned sft as an (user-steerable) alternative to rlhf.
\newblock \emph{arXiv preprint arXiv:2310.05344}.

\bibitem[{Hendrycks et~al.(2020)Hendrycks, Burns, Basart, Zou, Mazeika, Song, and Steinhardt}]{hendrycks2020measuring}
Dan Hendrycks, Collin Burns, Steven Basart, Andy Zou, Mantas Mazeika, Dawn Song, and Jacob Steinhardt. 2020.
\newblock Measuring massive multitask language understanding.
\newblock \emph{arXiv preprint arXiv:2009.03300}.

\bibitem[{Hoffmann et~al.(2022)Hoffmann, Borgeaud, Mensch, Buchatskaya, Cai, Rutherford, Casas, Hendricks, Welbl, Clark et~al.}]{hoffmann2022training}
Jordan Hoffmann, Sebastian Borgeaud, Arthur Mensch, Elena Buchatskaya, Trevor Cai, Eliza Rutherford, Diego de~Las Casas, Lisa~Anne Hendricks, Johannes Welbl, Aidan Clark, et~al. 2022.
\newblock Training compute-optimal large language models.
\newblock \emph{arXiv preprint arXiv:2203.15556}.

\bibitem[{Hu et~al.(2021)Hu, Shen, Wallis, Allen-Zhu, Li, Wang, Wang, and Chen}]{hu2021lora}
Edward~J Hu, Yelong Shen, Phillip Wallis, Zeyuan Allen-Zhu, Yuanzhi Li, Shean Wang, Lu~Wang, and Weizhu Chen. 2021.
\newblock Lora: Low-rank adaptation of large language models.
\newblock \emph{arXiv preprint arXiv:2106.09685}.

\bibitem[{Jiang et~al.(2023)Jiang, Sablayrolles, Mensch, Bamford, Chaplot, Casas, Bressand, Lengyel, Lample, Saulnier et~al.}]{jiang2023mistral}
Albert~Q Jiang, Alexandre Sablayrolles, Arthur Mensch, Chris Bamford, Devendra~Singh Chaplot, Diego de~las Casas, Florian Bressand, Gianna Lengyel, Guillaume Lample, Lucile Saulnier, et~al. 2023.
\newblock Mistral 7b.
\newblock \emph{arXiv preprint arXiv:2310.06825}.

\bibitem[{K{\"o}pf et~al.(2023)K{\"o}pf, Kilcher, von R{\"u}tte, Anagnostidis, Tam, Stevens, Barhoum, Duc, Stanley, Nagyfi et~al.}]{kopf2023openassistant}
Andreas K{\"o}pf, Yannic Kilcher, Dimitri von R{\"u}tte, Sotiris Anagnostidis, Zhi-Rui Tam, Keith Stevens, Abdullah Barhoum, Nguyen~Minh Duc, Oliver Stanley, Rich{\'a}rd Nagyfi, et~al. 2023.
\newblock Openassistant conversations--democratizing large language model alignment.
\newblock \emph{arXiv preprint arXiv:2304.07327}.

\bibitem[{Lin et~al.(2021)Lin, Hilton, and Evans}]{lin2021truthfulqa}
Stephanie Lin, Jacob Hilton, and Owain Evans. 2021.
\newblock Truthfulqa: Measuring how models mimic human falsehoods.
\newblock \emph{arXiv preprint arXiv:2109.07958}.

\bibitem[{Luukkonen et~al.(2023)Luukkonen, Komulainen, Luoma, Eskelinen, Kanerva, Kupari, Ginter, Laippala, Muennighoff, Piktus et~al.}]{luukkonen2023fingpt}
Risto Luukkonen, Ville Komulainen, Jouni Luoma, Anni Eskelinen, Jenna Kanerva, Hanna-Mari Kupari, Filip Ginter, Veronika Laippala, Niklas Muennighoff, Aleksandra Piktus, et~al. 2023.
\newblock Fingpt: Large generative models for a small language.
\newblock \emph{arXiv preprint arXiv:2311.05640}.

\bibitem[{Nguyen et~al.(2023)Nguyen, Van~Nguyen, Lai, Man, Ngo, Dernoncourt, Rossi, and Nguyen}]{nguyen2023culturax}
Thuat Nguyen, Chien Van~Nguyen, Viet~Dac Lai, Hieu Man, Nghia~Trung Ngo, Franck Dernoncourt, Ryan~A Rossi, and Thien~Huu Nguyen. 2023.
\newblock Culturax: A cleaned, enormous, and multilingual dataset for large language models in 167 languages.
\newblock \emph{arXiv preprint arXiv:2309.09400}.

\bibitem[{Radford et~al.(2018)Radford, Narasimhan, Salimans, Sutskever et~al.}]{radford2018improving}
Alec Radford, Karthik Narasimhan, Tim Salimans, Ilya Sutskever, et~al. 2018.
\newblock Improving language understanding by generative pre-training.

\bibitem[{Radford et~al.(2019)Radford, Wu, Child, Luan, Amodei, Sutskever et~al.}]{radford2019language}
Alec Radford, Jeffrey Wu, Rewon Child, David Luan, Dario Amodei, Ilya Sutskever, et~al. 2019.
\newblock Language models are unsupervised multitask learners.
\newblock \emph{OpenAI blog}, 1(8):9.

\bibitem[{Rafailov et~al.(2024)Rafailov, Sharma, Mitchell, Manning, Ermon, and Finn}]{rafailov2024direct}
Rafael Rafailov, Archit Sharma, Eric Mitchell, Christopher~D Manning, Stefano Ermon, and Chelsea Finn. 2024.
\newblock Direct preference optimization: Your language model is secretly a reward model.
\newblock \emph{Advances in Neural Information Processing Systems}, 36.

\bibitem[{Sakaguchi et~al.(2021)Sakaguchi, Bras, Bhagavatula, and Choi}]{sakaguchi2021winogrande}
Keisuke Sakaguchi, Ronan~Le Bras, Chandra Bhagavatula, and Yejin Choi. 2021.
\newblock Winogrande: An adversarial winograd schema challenge at scale.
\newblock \emph{Communications of the ACM}, 64(9):99--106.

\bibitem[{Taori et~al.(2023)Taori, Gulrajani, Zhang, Dubois, Li, Guestrin, Liang, and Hashimoto}]{taori2023alpaca}
Rohan Taori, Ishaan Gulrajani, Tianyi Zhang, Yann Dubois, Xuechen Li, Carlos Guestrin, Percy Liang, and Tatsunori~B Hashimoto. 2023.
\newblock Alpaca: A strong, replicable instruction-following model.
\newblock \emph{Stanford Center for Research on Foundation Models. https://crfm. stanford. edu/2023/03/13/alpaca. html}, 3(6):7.

\bibitem[{Touvron et~al.(2023)Touvron, Martin, Stone, Albert, Almahairi, Babaei, Bashlykov, Batra, Bhargava, Bhosale et~al.}]{touvron2023llama}
Hugo Touvron, Louis Martin, Kevin Stone, Peter Albert, Amjad Almahairi, Yasmine Babaei, Nikolay Bashlykov, Soumya Batra, Prajjwal Bhargava, Shruti Bhosale, et~al. 2023.
\newblock Llama 2: Open foundation and fine-tuned chat models.
\newblock \emph{arXiv preprint arXiv:2307.09288}.

\bibitem[{Vaswani et~al.(2017)Vaswani, Shazeer, Parmar, Uszkoreit, Jones, Gomez, Kaiser, and Polosukhin}]{vaswani2017attention}
Ashish Vaswani, Noam Shazeer, Niki Parmar, Jakob Uszkoreit, Llion Jones, Aidan~N Gomez, {\L}ukasz Kaiser, and Illia Polosukhin. 2017.
\newblock Attention is all you need.
\newblock \emph{Advances in neural information processing systems}, 30.

\bibitem[{Wang(2023)}]{wang2023mediagpt}
Zhonghao Wang. 2023.
\newblock Mediagpt: A large language model target chinese media.
\newblock \emph{arXiv preprint arXiv:2307.10930}.

\bibitem[{Xue et~al.(2021)Xue, Constant, Roberts, Kale, Al-Rfou, Siddhant, Barua, and Raffel}]{xue-etal-2021-mt5}
Linting Xue, Noah Constant, Adam Roberts, Mihir Kale, Rami Al-Rfou, Aditya Siddhant, Aditya Barua, and Colin Raffel. 2021.
\newblock \href {https://doi.org/10.18653/v1/2021.naacl-main.41} {m{T}5: A massively multilingual pre-trained text-to-text transformer}.
\newblock In \emph{Proceedings of the 2021 Conference of the North American Chapter of the Association for Computational Linguistics: Human Language Technologies}, pages 483--498, Online. Association for Computational Linguistics.

\bibitem[{Zellers et~al.(2019)Zellers, Holtzman, Bisk, Farhadi, and Choi}]{zellers2019hellaswag}
Rowan Zellers, Ari Holtzman, Yonatan Bisk, Ali Farhadi, and Yejin Choi. 2019.
\newblock Hellaswag: Can a machine really finish your sentence?
\newblock \emph{arXiv preprint arXiv:1905.07830}.

\bibitem[{Zheng et~al.(2024)Zheng, Chiang, Sheng, Zhuang, Wu, Zhuang, Lin, Li, Li, Xing et~al.}]{zheng2024judging}
Lianmin Zheng, Wei-Lin Chiang, Ying Sheng, Siyuan Zhuang, Zhanghao Wu, Yonghao Zhuang, Zi~Lin, Zhuohan Li, Dacheng Li, Eric Xing, et~al. 2024.
\newblock Judging llm-as-a-judge with mt-bench and chatbot arena.
\newblock \emph{Advances in Neural Information Processing Systems}, 36.

\bibitem[{Zheng et~al.(2023)Zheng, Chiang, Sheng, Zhuang, Wu, Zhuang, Lin, Li, Li, Xing, Zhang, Gonzalez, and Stoica}]{zheng2023judging}
Lianmin Zheng, Wei-Lin Chiang, Ying Sheng, Siyuan Zhuang, Zhanghao Wu, Yonghao Zhuang, Zi~Lin, Zhuohan Li, Dacheng Li, Eric~P. Xing, Hao Zhang, Joseph~E. Gonzalez, and Ion Stoica. 2023.
\newblock \href {http://arxiv.org/abs/2306.05685} {Judging llm-as-a-judge with mt-bench and chatbot arena}.

\bibitem[{Zhong et~al.(2023)Zhong, Cui, Guo, Liang, Lu, Wang, Saied, Chen, and Duan}]{zhong2023agieval}
Wanjun Zhong, Ruixiang Cui, Yiduo Guo, Yaobo Liang, Shuai Lu, Yanlin Wang, Amin Saied, Weizhu Chen, and Nan Duan. 2023.
\newblock Agieval: A human-centric benchmark for evaluating foundation models.
\newblock \emph{arXiv preprint arXiv:2304.06364}.

\bibitem[{Ziegler et~al.(2019)Ziegler, Stiennon, Wu, Brown, Radford, Amodei, Christiano, and Irving}]{ziegler2019fine}
Daniel~M Ziegler, Nisan Stiennon, Jeffrey Wu, Tom~B Brown, Alec Radford, Dario Amodei, Paul Christiano, and Geoffrey Irving. 2019.
\newblock Fine-tuning language models from human preferences.
\newblock \emph{arXiv preprint arXiv:1909.08593}.

\end{thebibliography}

\newpage
\appendix
\section{Detailed Results}
\label{sec:detailed_results}

In this section we present detailed results for Romanian versions of ARC-Challenge (Table~\ref{tab:arc_detailed}), MMLU (Table~\ref{tab:mmlu_detailed}), Winogrande (Table~\ref{tab:winogrande_detailed}), HellaSwag (Table~\ref{tab:hellaswag_detailed}) and GSM8k (Table~\ref{tab:gsm8k_detailed}).
ARC, HellaSwag, and Winogrande benchmarks are used to assess a model's reasoning capabilities, as ARC contains science exam questions and HellaSwag and Winogrande consist of real-world scenarios framed as text completion and fill-in-a-blank tasks, respectively. MMLU benchmark covers 57 subjects ranging from STEM to social sciences, humanities, and others, with difficulty ranging from elementary to professional level testing both the problem-solving capabilities and world knowledge of a model. TruthfulQA is a benchmark devised to measure whether a model is truthful in answering questions, while GSM8k consists of school math problems and is used for the evaluation of multi-step mathematical reasoning. While ARC, HellaSwag, Winogrande, MMLU, and TruthfulQA are posed as multi-choice tasks (i.e., the model has to choose the most probable answer), GSM8k is framed as a generation task, in which the model has to generate the solution. Finally, the results represent in all tables the percentage of cases in which the model picks or generates the correct solution.

\begin{table*}[hb]
\centering
\begin{tabular}{lccccccc}
\hline
\textbf{Model} & \textbf{ARC-0} &\textbf{ARC-1} &\textbf{ARC-3} &\textbf{ARC-5} &\textbf{ARC-10} & 
\textbf{ARC-25} &\textbf{ARC Avg} \\
\hline
\multicolumn{8}{c}{\textbf{\textit{Existing Models}}}\\
\hline
\multicolumn{8}{c}{\textit{For Romanian}}\\
Okapi & 36.08 & 36.93 & 37.02 & 36.85 & 36.85 & 37.53 & 36.88\\
Andrei481/Llama-2-7b-Romanian & 32.13 & 35.22 & 35.73 & 36.50 & 36.42 & 36.93 & 35.49\\
Andrei481/Mistral-7B-v0.1-Romanian & 36.08 & 37.36 & 38.56 & 39.07 & 39.25 & 36.50 & \textbf{37.80}\\
\hline
\multicolumn{8}{c}{\textit{General}}\\
Llama-2-7b & 31.36 & 33.93 & 34.10 & 34.36 & 34.45 & 34.88 & 33.85\\
Llama-2-7b-chat & 32.73 & 35.22 & 34.53 & 35.90 & 35.13 & 35.99 & 34.92\\
Mistral-7B-v0.1 & 38.64 & 41.05 & 40.62 & 42.50 & 41.65 & 42.84 & 41.22\\
Mistral-7B-Instruct-v0.2 & 40.02 & 43.02 & 43.10 & 44.47 & 43.79 & 44.13 & \textbf{43.09}\\
\hline\hline
\multicolumn{8}{c}{\textbf{\textit{Romanian LLMs}}}\\
\textit{RoLlama2-7b-Base} & 33.33 & 35.30 & 35.82 & 37.28 & 36.76 & 36.50 & 35.83\\
\textit{RoLlama2-7b-Instruct} & 40.27 & 39.16 & 40.36 & 40.27 & 41.30 & 40.79 & 40.36\\
\textit{RoLlama2-7b-Chat} & 36.85 & 38.22 & 38.39 & 38.39 & 38.99 & 38.90 & 38.29 \\
\textit{RoMistral-7b-Instruct} & 47.39	& 48.67 &	50.30 &	51.41 &	52.27 &	52.27 &	\textbf{50.39}\\

\hline
\end{tabular}
\caption{ARC performance under various few-shot settings.}
\label{tab:arc_detailed}
\end{table*}

\begin{table*}
\centering
\begin{tabular}{lccccc}
\hline
\textbf{Model} & \textbf{MMLU-0} &\textbf{MMLU-1} &\textbf{MMLU-3} &\textbf{MMLU-5} &\textbf{MMLU Avg} \\
\hline
\multicolumn{6}{c}{\textbf{\textit{Existing Models}}}\\
\hline
\multicolumn{6}{c}{\textit{For Romanian}}\\
Okapi & 30.81 & 30.48 & 31.51 & 31.87 & 31.17 \\
Andrei481/Llama-2-7b-Romanian & 28.88 & 30.03 & 31.97 & 32.55 & 30.86\\
Andrei481/Mistral-7B-v0.1-Romanian & 30.91 & 34.13 & 34.69 & 33.61 & \textbf{33.34}\\
\hline
\multicolumn{6}{c}{\textit{General}}\\
Llama-2-7b & 27.93 & 30.62 & 32.16 & 32.99 & 30.93\\
Llama-2-7b-chat & 30.47 & 31.53 & 33.41 & 34.06 & 32.37\\
Mistral-7B-v0.1 & 35.23 & 41.62 & 42.98 & 43.52 & 40.84\\
Mistral-7B-Instruct-v0.2 & 41.63 & 44.81 & 46.65 & 46.40 & \textbf{44.87}\\
\hline\hline
\multicolumn{6}{c}{\textbf{\textit{Romanian LLMs}}}\\
\textit{RoLlama2-7b-Base} & 28.03 & 30.14 & 31.65 & 32.04 & 30.47\\
\textit{RoLlama2-7b-Instruct} & 33.82 & 37.70 & 38.70 & 39.42 & 37.41\\
\textit{RoLlama2-7b-Chat} & 30.54 & 35.64 & 37.41 & 37.47 & 35.27\\
\textit{RoMistral-7b-Instruct} & 49.95 & 50.32 & 53.10 & 53.19 & \textbf{51.64}\\

\hline
\end{tabular}
\caption{MMLU performance under various few-shot settings.}
\label{tab:mmlu_detailed}
\end{table*}

\begin{table*}
\centering
\begin{tabular}{lccccc}
\hline
\textbf{Model} & \textbf{Wino-0} &\textbf{Wino-1} &\textbf{Wino-3} &\textbf{Wino-5} &\textbf{Wino Avg} \\
\hline
\multicolumn{6}{c}{\textbf{\textit{Existing Models}}}\\
\hline
\multicolumn{6}{c}{\textit{For Romanian}}\\
Okapi & 55.80 & 56.67 & 56.12 & 56.51 & 56.28 \\
Andrei481/Llama-2-7b-Romanian & 55.80 & 56.04 & 55.96 & 56.12 & 55.98\\
Andrei481/Mistral-7B-v0.1-Romanian & 58.48 & 54.93 & 55.80 & 56.27 & \textbf{56.37}\\
\hline
\multicolumn{6}{c}{\textit{General}}\\
Llama-2-7b & 55.09 & 57.06 & 57.22 & 56.35 & 56.43\\
Llama-2-7b-chat & 53.75 & 55.01 & 54.06 & 54.22 & 54.26\\
Mistral-7B-v0.1 & 60.06 & 60.38 & 60.30 & 60.06 & \textbf{60.20}\\
Mistral-7B-Instruct-v0.2 & 60.30 & 59.43 & 58.17 & 59.12 & 59.26\\
\hline\hline
\multicolumn{6}{c}{\textbf{\textit{Romanian LLMs}}}\\
\textit{RoLlama2-7b-Base} & 59.98 & 59.67 & 60.22 & 60.77 & 60.16\\
\textit{RoLlama2-7b-Instruct} & 70.96 & 69.38 & 68.90 & 69.06 & \textbf{69.58}\\
\textit{RoLlama2-7b-Chat} & 65.59 & 65.11 & 66.14 & 64.17 & 65.25\\
\textit{RoMistral-7b-Instruct} & 65.43 & 67.72 & 66.69 & 66.93 & 66.69\\
\hline
\end{tabular}
\caption{Winogrande performance under various few-shot settings.}
\label{tab:winogrande_detailed}
\end{table*}

\begin{table*}
\centering
\begin{tabular}{lcccccc}
\hline
\textbf{Model} & \textbf{HS-0} &\textbf{HS-1} &\textbf{HS-3} &\textbf{HS-5} &\textbf{HS-10} & \textbf{HS Avg} \\
\hline
\multicolumn{7}{c}{\textbf{\textit{Existing Models}}}\\
\hline
\multicolumn{7}{c}{\textit{For Romanian}}\\
Okapi & 49.11 & 48.31 & 49.06 & 48.87 & 49.20 & 48.91 \\
Andrei481/Llama-2-7b-Romanian & 47.76 & 47.58 & 48.11 & 48.24 & 48.27 & 47.99\\
Andrei481/Mistral-7B-v0.1-Romanian & 51.56 &	48.27&	49.51&	48.99	&48.77	&\textbf{49.42}\\
\hline
\multicolumn{7}{c}{\textit{General}}\\
Llama-2-7b & 46.74&	46.44&	46.71&	47.33	&47.70&	46.98\\
Llama-2-7b-chat & 44.47&	43.96&	44.16&	45.01	&45.00&	44.52\\
Mistral-7B-v0.1 & 52.72	&53.02&	53.69&	54.25	&54.49	&53.63\\
Mistral-7B-Instruct-v0.2 & 54.44  &	53.60	 &53.96	 &54.07	 &54.51	 &\textbf{54.12}\\
\hline\hline
\multicolumn{7}{c}{\textbf{\textit{Romanian LLMs}}}\\
\textit{RoLlama2-7b-Base} & 54.98& 55.11& 	55.80	& 55.74& 	55.97& 	55.52\\
\textit{RoLlama2-7b-Instruct} & 54.25&	55.63	&56.08	&56.05	&56.17&	55.64\\
\textit{RoLlama2-7b-Chat} & 56.20 & 55.98 & 56.41 & 56.69	&56.99	& 56.45\\
\textit{RoMistral-7b-Instruct} & 59.98 & 59.45 & 60.16 & 60.55 & 61.06 & \textbf{60.24}\\

\hline
\end{tabular}
\caption{HellaSwag performance under various few-shot settings.}
\label{tab:hellaswag_detailed}
\end{table*}

\begin{table*}
\centering
\begin{tabular}{lcccc}
\hline
\textbf{Model} & \textbf{GSM8k-1} &\textbf{GSM8k-3} &\textbf{GSM8k-5} &\textbf{GSM8k Avg} \\
\hline
\multicolumn{5}{c}{\textbf{\textit{Existing Models}}}\\
\hline
\multicolumn{5}{c}{\textit{For Romanian}}\\
Okapi & 0.68&	0.15&	0.45&	0.43\\
Andrei481/Llama-2-7b-Romanian & 0.76&	0.91	&0.83&	0.83\\
Andrei481/Mistral-7B-v0.1-Romanian & 0.99	&1.29&	0.68&	\textbf{0.99}\\
\hline
\multicolumn{5}{c}{\textit{General}}\\
Llama-2-7b & 0.61&	1.52&	1.97&	1.37\\
Llama-2-7b-chat & 1.21&	1.90	&3.03&	2.05\\
Mistral-7B-v0.1 & 6.14&	17.06	&18.27&	\textbf{13.82}\\
Mistral-7B-Instruct-v0.2 & 0.45	&14.40&	17.74	&10.86\\
\hline\hline
\multicolumn{5}{c}{\textbf{\textit{Romanian LLMs}}}\\
\textit{RoLlama2-7b-Vase} & 0.00	&2.27&	4.25	&2.17\\
\textit{RoLlama2-7b-Instruct} & 12.66	&19.56&	20.55	& 17.59\\
\textit{RoLlama2-7b-Chat} & 9.40 &	14.03 &	15.09 &	12.84\\
\textit{RoMistral-7b-Instruct} & 21.30 & 37.98 & 41.85 & \textbf{33.71}\\
\hline
\end{tabular}
\caption{GSM8k performance under various few-shot settings.}
\label{tab:gsm8k_detailed}
\end{table*}

\end{document}